\def\year{2019}\relax
\documentclass[letterpaper]{article} 
\usepackage{aaai19}  
\usepackage{times}  
\usepackage{helvet}  
\usepackage{courier}  
\usepackage[utf8]{inputenc} 
\usepackage[T1]{fontenc}    
\usepackage{hyperref}       
\usepackage{url}            
\usepackage{booktabs}       
\usepackage{amsfonts,bm,amsmath}       
\usepackage{nicefrac}       
\usepackage{microtype}      
\usepackage{natbib}
\usepackage{graphicx,float,subfigure}
\usepackage{epstopdf}
\usepackage{color}
\usepackage{lipsum}
\usepackage{caption}
\usepackage[labelfont=bf]{caption}
\usepackage{array}

\newcolumntype{P}[1]{>{\centering\arraybackslash}m{#1}}

\newcommand{\argmax}{\operatornamewithlimits{argmax}}
\makeatletter
  \newcommand\figcaption{\def\@captype{figure}\caption}
  \newcommand\tabcaption{\def\@captype{table}\caption}
\makeatother

\newcommand*{\boxedcolor}{red}
\makeatletter
\renewcommand{\boxed}[1]{\textcolor{\boxedcolor}{%
  \fbox{\normalcolor\m@th$\displaystyle#1$}}}
\makeatother

\begin{document}
%
\title{Towards Highly Accurate and Stable Face Alignment for High-Resolution Videos}
\author{
Ying Tai$^{\dag \ast}$ Yicong Liang$^{\dag \ast}$ Xiaoming Liu$^{\ddag}$ Lei Duan$^{\S}$ Jilin Li$^{\dag}$ Chengjie Wang$^{\dag}$ Feiyue Huang$^{\dag}$ Yu Chen$^{\pounds}$ \\
$^{\dag}$Youtu Lab, Tencent ~~~
$^{\ddag}$Michigan State University ~~~ \\
$^{\S}$Fudan University ~~~
$^{\pounds}$Nanjing University of Science and Technology ~~~ \\
{\tt\small $^{\dag}$\{yingtai, easonliang, jerolinli, jasoncjwang, garyhuang\}@tencent.com}\\
{\tt\small $^{\ddag}$liuxm@cse.msu.edu, $^{\S}$$15307130193$@fudan.edu.cn, $^{\pounds}$chenyu1523@gmail.com} \\
{\url{https://github.com/tyshiwo/FHR_alignment}}
}

\maketitle
{
\renewcommand{\thefootnote}{\fnsymbol{footnote}}
\footnotetext{$\ast$ indicates equal contributions.}
}

\begin{abstract}
  In recent years, heatmap regression based models have shown their effectiveness in face alignment and pose estimation.
  However, Conventional Heatmap Regression (CHR) is not accurate nor stable when dealing with high-resolution facial videos, since it finds the maximum activated location in heatmaps which are generated from rounding coordinates,
  and thus leads to quantization errors when scaling back to the original high-resolution space.
  In this paper, we propose a Fractional Heatmap Regression (FHR) for high-resolution video-based face alignment.
  The proposed FHR can accurately estimate the fractional part according to the $2$D Gaussian function by sampling three points in heatmaps.
  To further stabilize the landmarks among continuous video frames while maintaining the precise at the same time, we propose a novel stabilization loss that contains two terms to address time delay and non-smooth issues, respectively.
  Experiments on $300$W, $300$-VW and Talking Face datasets clearly demonstrate that the proposed method is more accurate and stable than the state-of-the-art models.
\end{abstract}

\section{Introduction} \label{sec:1}
\noindent Face alignment aims to estimate a set of facial landmarks given a face image or video sequence.
It is a classic computer vision problem that has attributed to many advanced machine learning algorithms~\cite{Fan2018alignment,Bulat2017alignment,trigeorgis2016mnemonic,Peng2015PIEFA,Peng2016REDNet,kowalski2017deep,chen2017adversarial,Liu2017TSTN,Hu2018alignment}.
Nowadays, with the rapid development of consumer hardwares (e.g., mobile phones, digital cameras), High-Resolution (HR) video sequences can be easily collected.
Estimating facial landmarks on such high-resolution facial data has tremendous applications, e.g., face makeup~\cite{chen2017makeup}, editing with special effects~\cite{Korshunova2017swap} in live broadcast videos.
However, most existing face alinement methods work on faces with medium image resolutions~\cite{chen2017adversarial,Bulat2017alignment,Peng2016REDNet,Liu2017TSTN}.
Therefore, developing face alignment algorithms for {\it high-resolution videos} is at the core of this paper.

\begin{figure}
  \centering
  \includegraphics[width=0.48\textwidth]{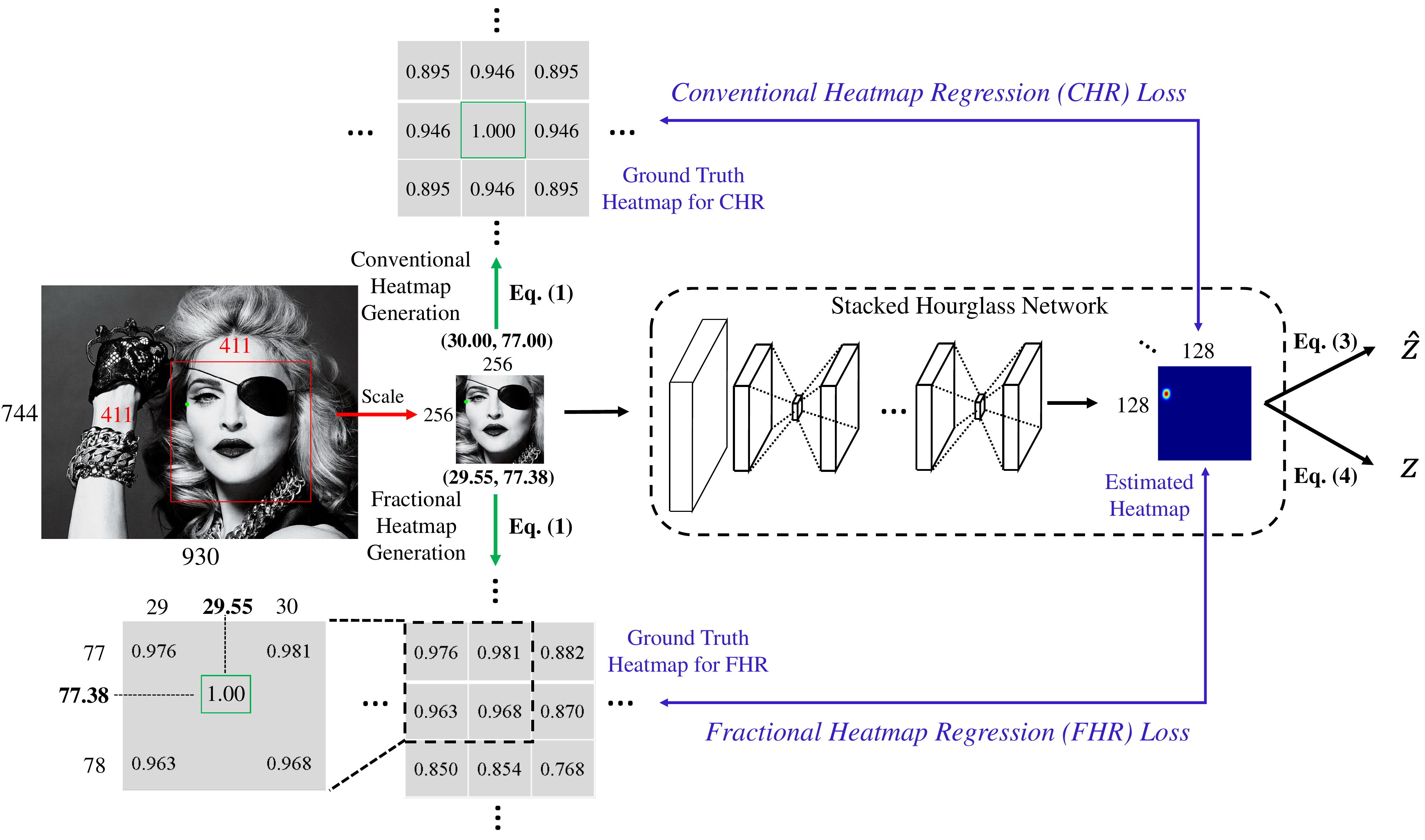} \\
  \vspace{-3mm}
  \caption{\small Comparisons between fractional regression heatmap and conventional heatmap regression. Our method differs conventional one in two aspects: 1) the ground truth heatmap for FHR maintains the precision of fractional coordinate, while the conventional one \textit{discards} (e.g., from $29.55$ to $30.00$, $77.38$ to $77.00$); and 2) three sampled points on the heatmap analytically computes the fractional peak location of the heatmap (Eq.~\ref{eq3.1.3}), while the conventional one only finds the maximum activated location~(Eq.~\ref{eq3.1.2}) that loses the fractional part and thus leads to \textit{quantization error}. }
  \label{fig:Network}
\end{figure}

To this end, we propose an accurate and stable algorithm for high-resolution video-based face alignment, named Fractional Heatmap Regression (FHR).
It is well known that heatmap regression have shown its effectiveness in landmark estimation tasks~\cite{chen2017adversarial,Newell2016SHN,CT-FSRNet-2018}.
However, Conventional Heatmap Regression (CHR) is not accurate nor stable when dealing with high-resolution facial images, since it finds the maximum activated location in heatmaps which are generated from \textit{rounding} coordinates,
and thus leads to quantization errors as the heatmap resolution is much lower than the input image resolution (e.g., $128$ vs. $930$ shown in Fig.~\ref{fig:Network}) due to the \textit{scaling operation}.
To address this problem, we propose a novel transformation between heatmaps and coordinates, which not only preserves the fractional part when generating heatmaps from the coordinates, but also accurately estimate the fractional part according to the $2$D Gaussian function by sampling three points in heatmaps.

Using our proposed FHR, we can estimate more accurate landmarks compared to the conventional heatmap regression model, and achieve state-of-the-art performance on popular video benchmarks: Talking Face~\cite{TF2014} and $300$-VW datasets~\cite{Shen2015fa300vw} compared to recent video-based face alignment models~\cite{Liu2017TSTN,Peng2016REDNet}. 
However, real-world applications such as face makeup in videos often demands extremely high {\it stability}, since the makeup jumps if the estimated landmarks oscillate between consecutive frames, which negatively impacts the user's experience.
To make the sequential estimations as stable as possible, we further develop a novel stabilization algorithm on the landmarks estimated by FHR, which contains two terms, a regularization term ($\mathcal{L}_{reg}$) and a temporal coherence term ($\mathcal{L}_{tm}$), to address two common difficulties: time delay and non-smooth problems, respectively.
Specifically, $\mathcal{L}_{reg}$ combines traditional Euclidean loss and a novel loss account for time delay; $\mathcal{L}_{tm}$ generalizes the temporal coherence loss in~\cite{Cao2014DDE} to better handle nonlinear movement of facial landmark.

In summary, the main contributions of this paper are:
\begin{itemize}
\item A novel Fractional Heatmap Regression method for high-resolution video based face alignment that leverages $2$D Gaussian prior to preserve the fractional part of points. 
\item A novel stabilization algorithm that addresses time delay and non-smooth problems among continuous video frames is proposed.
\item State-of-the-art performance, both in accuracy and stability, on the benchmarks of $300$W~\cite{Sagonas2013align}, $300$-VW~\cite{Shen2015fa300vw} and Talking Face~\cite{TF2014} datasets.
\end{itemize}

\section{Related Work} \label{sec:2}

\paragraph{Heatmap Regression}
Heatmap regression is one of the most widely used approaches for landmark localization tasks, which estimates a set of heatmaps rather than coordinates.
Stacked Hourglass Networks (SHN) are popular architectures in heatmap regression, which have symmetric topology that capture and consolidate information across all scales of the image.
Newell et al.~\cite{Newell2016SHN} proposed SHN for $2$D human pose estimation, which achieved remarkable results even for very challenging datasets~\cite{Andriluka2014pose}.
With the hourglass structure, Chu et al.~\cite{Chu2017pose} introduced multi-context attention mechanism into Convolutional Neural Networks (CNN).
Apart from applications in human pose estimation, there are also several heatmap regression based models for face alignment.
Chen et al.~\cite{chen2017adversarial} proposed a structure-aware fully convolutional network to implicitly model the priors during training.
Bulat et al.~\cite{Bulat2017alignment} built a powerful CNN for face alignment based on the hourglass network and a hierarchical, parallel and multi-scale block. 

However, all existing models drops the fractional part of coordinates during the transformation between heatmaps and points, which brings quantization errors to high-resolution facial images.
On the contrary, our proposed FHR can accurately estimate the fractional part according to the $2$D Gaussian function by sampling three points in heatmaps, and thus achieves more accurate alignment.

\vspace{-2mm}
\paragraph{Video-based Face Alignment}
Video-based face alignment estimates facial landmarks in video sequences~\cite{liu2010video}.
Early methods~\cite{Black1995alignment,Shen2015fa300vw} used incremental learning to predict landmarks on still frames in a tracking-by-detection manner.
To address the issue that generic methods are sensitive to initializations, Peng et al.~\cite{Peng2015PIEFA} exploited incremental learning for personalized ensemble alignment, which samples multiple initial shapes to achieve image congealing within one frame.
To explicitly model the temporal dependency~\cite{Oh2015ACV} of landmarks across frames, the authors~\cite{Peng2016REDNet} further incorporated a sequence of spatial and temporal recurrences for sequential face alignment in videos.
Recently, Liu et al.~\cite{Liu2017TSTN} proposed a Two-Stream Transformer Networks (TSTN) approach, which captures the complementary information of both the spatial appearance on still frames and the temporal consistency across frames.
Different from~\cite{Peng2016REDNet,Liu2017TSTN} that require temporal landmark labels across frames, our proposed method achieves state-of-the-art accuracy only by making full use of the spatial appearance on still frames, which is able to remedy the problem that \textit{labeled sequential data are very limited}.

Apart from the accuracy of landmarks, stabilization is also a key metric to video-based alignment.
Typically, two terms~\cite{Cao2014DDE} are adopted for stabilization, where a regularization term drives the optimized results to be more expressive, and a temporal coherence term drives the results to be more stable and smooth.
However, the existing stabilization algorithm is sensitive to time delay and nonlinear movements.
Our proposed algorithm takes these into account and thus are  overall more robust. 

\begin{figure*}
  \centering
  \subfigure[]{
    \label{fig:SHN_FSHN:a} 
    \includegraphics[width=0.465\textwidth]{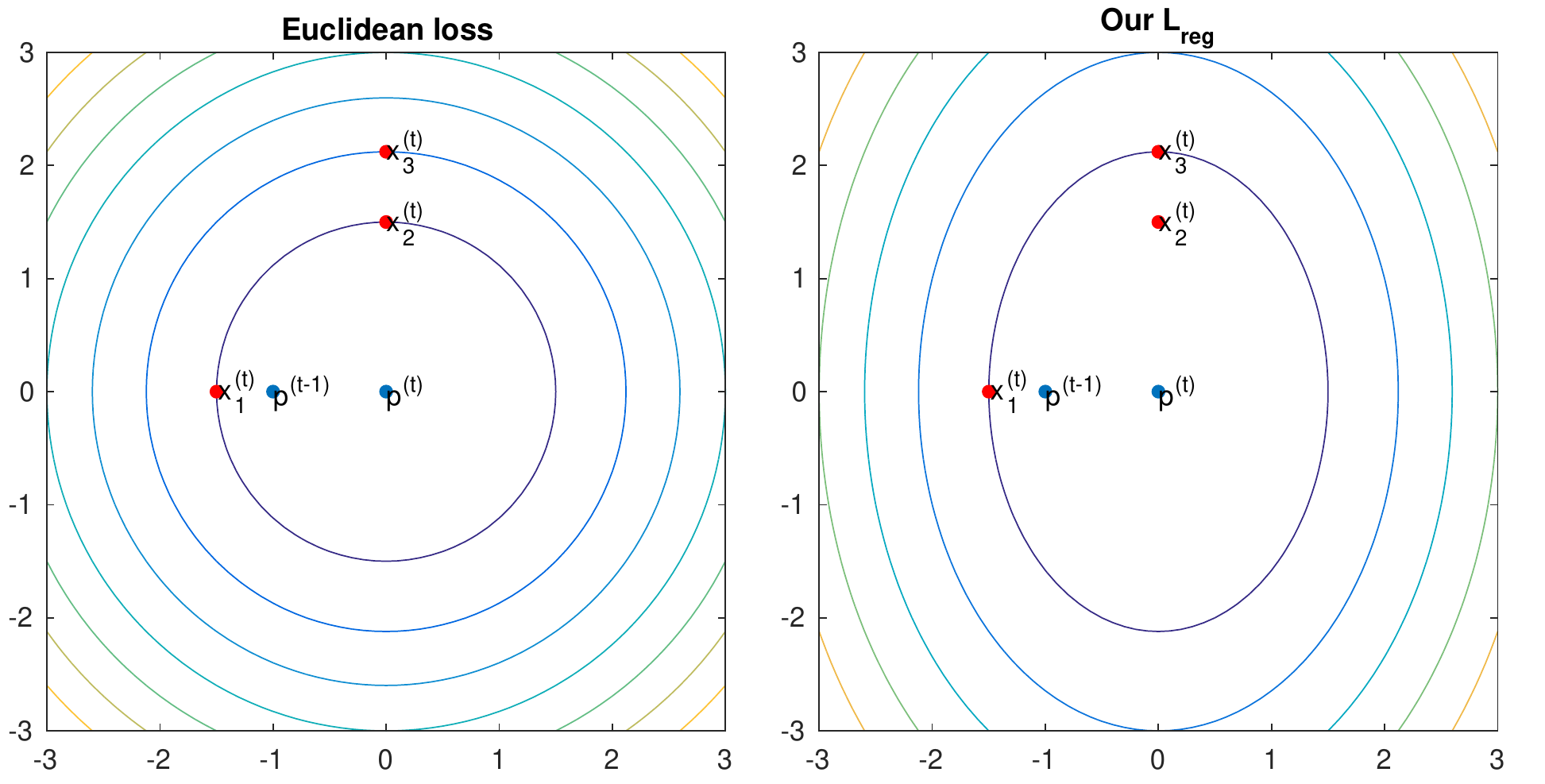}}
  \vspace{1in}
  \subfigure[]{
    \label{fig:SHN_FSHN:b} 
    \includegraphics[width=0.465\textwidth]{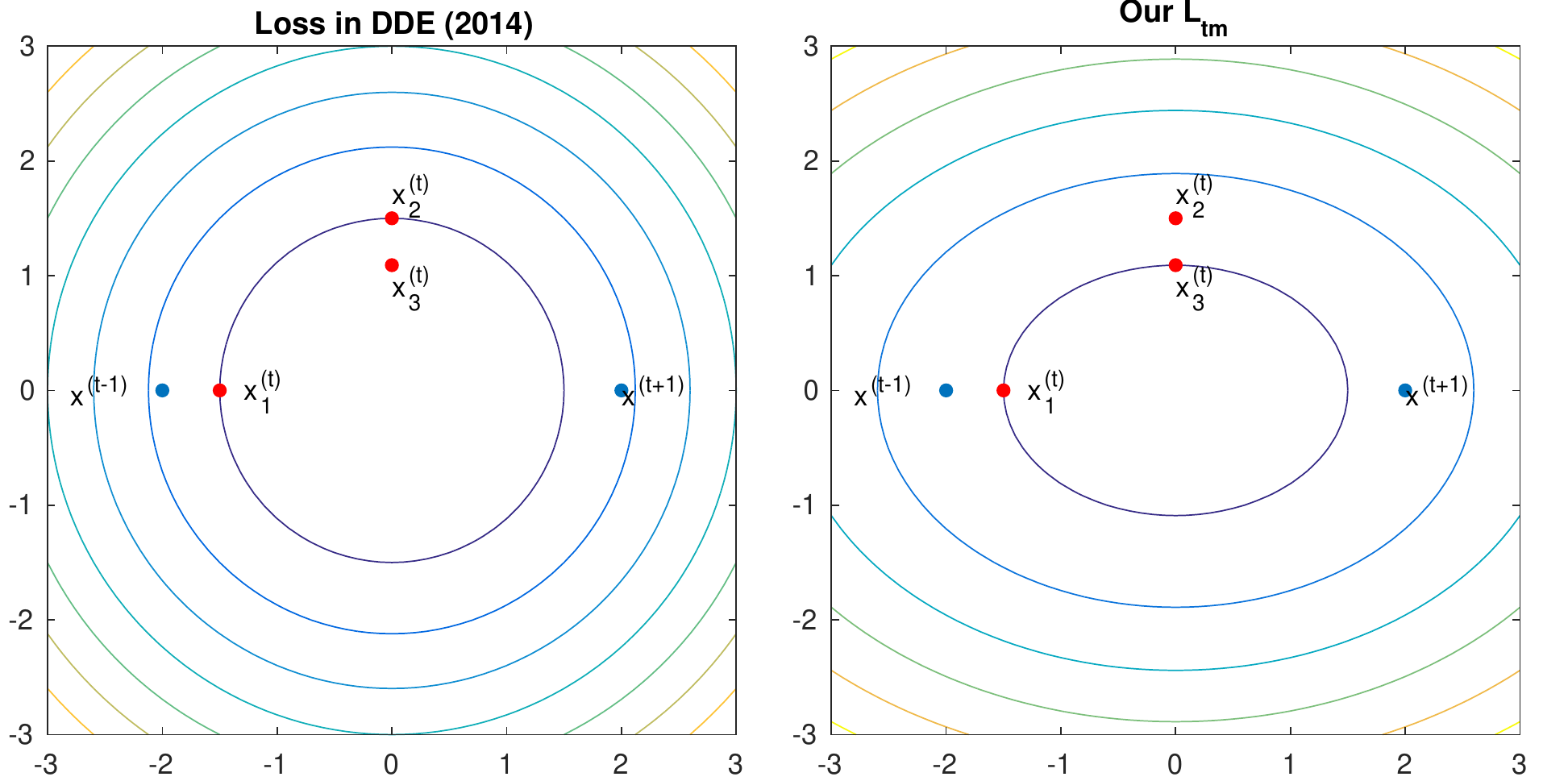}}
  \vspace{-30mm}
  \caption{\small (a) Histograms of  the Euclidean loss (left) and our $\mathcal{L}_{reg}$ (right). $\mathbf{p}^{(t)}$ and $\mathbf{p}^{(t-1)}$ are ground truths of frame $t$ and $t-1$, respectively. $\mathbf{x}_1^{(t)}$ has the same Euclidean loss as $\mathbf{x}_2^{(t)}$. However, since $\mathbf{x}_1^{(t)}$ lies on the line $\overline{\mathbf{p}^{(t)}\mathbf{p}^{(t-1)}}$, it indicates a loss caused by time delay, which is more likely to happen in the stabilization process.
  Thus our model prefers to assign it a larger loss (equal to $\mathbf{x}_3^{(t)}$).
  (b) Histograms of the loss in~\cite{Cao2014DDE} (left) and our $\mathcal{L}_{tm}$ (right). $\mathbf{x}^{(t-1)}$ and $\mathbf{x}^{(t+1)}$ are stabilization outputs of frame $t-1$ and $t+1$, respectively.
  $\mathbf{x}_1^{(t)}$ has the same loss as $\mathbf{x}_2^{(t)}$ in~\cite{Cao2014DDE}.
  However, since $\mathbf{x}_1^{(t)}$ lies on the line $\overline{\mathbf{x}^{(t-1)}\mathbf{x}^{(t+1)}}$, we argue that movement trajectory $\protect\overrightarrow{{\mathbf{x}^{(t-1)}\mathbf{x}_1^{(t)}\mathbf{x}^{(t+1)}}}$ is more smooth than trajectory $\protect\overrightarrow{{\mathbf{x}^{(t-1)}\mathbf{x}_2^{(t)}\mathbf{x}^{(t+1)}}}$.
  Thus our model assigns it a smaller loss (equal to $\mathbf{x}_3^{(t)}$).
 Note that the short axis of the ellipse in (a) is $\overline{\mathbf{p}^{(t)}\mathbf{p}^{(t-1)}}$ while the long axis of the ellipse in (b) is $\overline{\mathbf{x}^{(t-1)}\mathbf{x}^{(t+1)}}$, thus the two terms are not in contradiction.
  }
  \label{fig:Stable_loss} 
\end{figure*}

\vspace{-1mm}
\section{The Proposed Approach} \label{sec:3}
In this section, we introduce the details of the proposed approach based on heatmap regression.
A key point of heatmap regression is the transformation between the heatmaps and coordinates.
Specifically, before model training, a \textit{pre-process step} is conducted to convert the coordinates to the heatmaps, which are used as the ground truth labels.
After estimating the heatmaps, a \textit{post-process step} is conducted to obtain the coordinates from the estimated heatmaps.
In this work, we propose a novel transformation between the heatmaps and coordinates that is different from the conventional heatmap regression, which is demonstrated to be simple yet effective.

\vspace{-1mm}
\subsection{Fractional Heatmap Regression} \label{sec:3.1}
As shown in Fig.~\ref{fig:Network}, conventional heatmap regression mainly generates the heatmaps from integral coordinates, despite that the ground truth coordinates are usually with fractions.
As a result, it causes quantization errors when scaling back to the original image resolution, since the heatmaps are usually of much lower resolution than the input image.
To address this problem, our proposed fractional heatmap regression generates ground truth heatmaps based on the intact ground truth coordinates (see Fig.~\ref{fig:Network}) as follows:
\begin{equation}
\label{eq3.1.0}
\footnotesize
    \mathbf{\bar{H}}_{(m)}(\mathbf{c}) = \exp\left(-\frac{1}{2\sigma^2}((c_x-c_{(m)x_0})^2+(c_y-c_{(m)y_0})^2)\right),
\end{equation}
where $\mathbf{c}=(c_x, c_y)\in \Omega$ represents the coordinate, $\Omega$ is the domain of the heatmap $\mathbf{H}$, $\sigma$ denotes the standard deviation and $(c_{(m)x_0},c_{(m)y_0})$ is the center of the $2$D Gaussian in the $m$th heatmap.

Denoting $\mathbf{I}$ an input image and $\mathcal{F}$ the deep alignment model, we can estimate the recovered heatmaps by
\begin{equation}
\label{eq3.1.1}
    \mathbf{\hat{H}} = \mathcal{F}(\mathbf{I}),
\end{equation}
where $ \mathbf{\hat{H}} =[\mathbf{\hat{H}}_{(1)}, ..., \mathbf{\hat{H}}_{(m)}, ..., \mathbf{\hat{H}}_{(M)}]$, and $M$ is the number of landmarks.
Given Eq.~(\ref{eq3.1.0}) and $\mathbf{\hat{H}}_{(m)}$, estimating the fractional coordinate $\mathbf{z}$ amounts to solving a binary quadratic equation, which has a closed-form solution as long as we can sample any \textit{three} non-zero points from the heatmap.
Specifically, we first obtain $\mathbf{\hat{z}}$ as the location $\mathbf{c}$ with the \textit{maximum likelihood}, as in conventional heatmap regression:
\begin{equation}
\label{eq3.1.2}
    \mathbf{\hat{z}}_{m} = \argmax_{\mathbf{c}}{\mathbf{\hat{H}}}_{(m)}(\mathbf{c}), \ m=1,...,M.
\end{equation}
Conventional heatmap regression directly takes $\mathbf{\hat{z}}_{m}$ as the output, which loses the fractional part.
In our method, we further sample another two points, e.g., $\mathbf{\hat{z}}^1_{m}=({\hat{z}}_{{(m)}x}+1,{\hat{z}}_{{(m)}y}),\mathbf{\hat{z}}^2_{m}=({\hat{z}}_{{(m)}x},{\hat{z}}_{{(m)}y}+1)$ near $\mathbf{\hat{z}}_{m}=({\hat{z}}_{{(m)}x},{\hat{z}}_{{(m)}y})$~\footnote{In case $\mathbf{\hat{z}}_{m}$ is located at the edge of the heatmap, we would sample the points in the opposite directions.}.
Let $h_{(m)}^1=\mathbf{\hat{H}}_{(m)}(\mathbf{\hat{z}}^1_{m})$, $h_{(m)}^2=\mathbf{\hat{H}}_{(m)}(\mathbf{\hat{z}}^2_{m})$ and $h_{(m)}=\mathbf{\hat{H}}_{(m)}(\mathbf{\hat{z}}_{m})$, we then estimate the \textit{fractional coordinate} $\mathbf{z}_{m}= ({z}_{(m)x},{z}_{(m)y})$ as follows:
\begin{equation}
\label{eq3.1.3}
\begin{aligned}
    z_{(m)x} = &\sigma^2(\ln{h_{(m)}^1} - \ln{h_{(m)}}) -  \\
    & \frac{1}{2}((\hat{z}_{{(m)}x})^2 - (\hat{z}_{{(m)}x}^1)^2 + (\hat{z}_{{(m)}y})^2 - (\hat{z}_{{(m)}y}^1)^2), \\
    z_{(m)y} = &\sigma^2(\ln{h_{(m)}^2} - \ln{h_{(m)}}) -  \\
    & \frac{1}{2}((\hat{z}_{{(m)}x})^2 - (\hat{z}_{{(m)}x}^2)^2 + (\hat{z}_{{(m)}y})^2 - (\hat{z}_{{(m)}y}^2)^2).
\end{aligned}
\end{equation}
It should be noted that our fractional heatmap regression is applicable to any heatmap based methods.
In this paper, we focus on face alignment, and adopt the stacked hourglass network~\cite{Newell2016SHN} as the alignment model $\mathcal{F}$ that minimizes the loss of $\|\mathbf{\bar{H}}-\mathbf{\hat{H}}\|^2$ across the entire training set, where $\mathbf{\bar{H}}=[\mathbf{\bar{H}}_{(1)}, ..., \mathbf{\bar{H}}_{(m)}, ..., \mathbf{\bar{H}}_{(M)}]$.

\vspace{-1mm}
\subsection{Stabilization Algorithm}
\label{sec:3.2}

We now introduce our stabilization algorithm for video-based alignment, which takes the alignment results of $\mathcal{F}$ in all the past frames as input, and outputs a more accurate and stable result for the current frame.

\subsubsection{Stabilization Model}
\label{sec:3.2.1}

We denote $\mathbf{z}^{(t)}$ as the output of $\mathcal{F}$ at frame $t$ and the stabilization model as $\mathcal{M}_\Theta$, which has parameters $\Theta$ to be optimized.
$\mathcal{M}_\Theta$ takes $\mathbf{z}^{(1)},\ldots,\mathbf{z}^{(t)}$ as input, and outputs the stabilized landmarks of frame $t$, which is denoted as $\mathbf{x}^{(t)}$. Therefore we have
\begin{equation}
\label{eq3.2.2.0}
    \mathbf{x}^{(t)} = \mathcal{M}_\Theta\left(\mathbf{z}^{(1)},\ldots,\mathbf{z}^{(t)}\right).
\end{equation}
Assume there are $V$ videos in the training set, and the $i$th video has $T_i$ frames.
For frame $t$ in the $i$th video, we denote its ground truth landmarks  as $\mathbf{p}_i^{(t)}$, the output of $\mathcal{F}$ as $\mathbf{z}_i^{(t)}$, and the stabilized output as $\mathbf{x}_i^{(t)}$ ($\mathbf{p}_i^{(t)}, \mathbf{z}_i^{(t)}, \mathbf{x}_i^{(t)} \in \mathbb{R}^{2M \times 1}$). 
Here, we have $\mathbf{x}_i^{(t)} = \mathcal{M}_\Theta\left(\mathbf{z}_i^{(1)},\ldots,\mathbf{z}_i^{(t)}\right)$.

Next, we present the specific form of $\mathcal{M}_\Theta$ as well as its parameters $\Theta$.
Our model follows a Bayesian framework.
Specifically, we model the prior distribution of $\mathbf{x}_i^{(t)}$ given $\mathbf{z}^{(1)}, \ldots, \mathbf{z}^{(t-1)}$ as a $K$-component Gaussian mixture:
\begin{equation}
\label{eq3.2.2.1}
    Pr\left(\mathbf{x}^{(t)}|\mathbf{z}^{(1)},\ldots,\mathbf{z}^{(t-1)}\right) = \sum_{k=1}^K\alpha_k\mathcal{N}\left(\mathbf{x}^{(t)};\mu_k^{(t)},\Sigma_k^{(t)}\right),
\end{equation}
where $Pr(\cdot)$ indicates the density function, $\mathcal{N}(\cdot; \mu,\Sigma)$ indicates a normal distribution with mean $\mu$ and covariance $\Sigma$.
We then model the likelihood of $\mathbf{z}^{(t)}$ given $\mathbf{x}^{(t)}$ as Gaussian
\begin{equation}
\label{eq3.2.2.2}
    Pr\left(\mathbf{z}^{(t)}|\mathbf{x}^{(t)}\right) = \mathcal{N}\left(\mathbf{z}^{(t)};\mathbf{x}^{(t)},\Sigma_{noise}\right),
\end{equation}
and use the Bayesian rule to obtain the most probable value of $\mathbf{x}^{(t)}$:
\begin{equation}
\tiny
\label{eq3.2.2.3}
\begin{aligned}
    \mathbf{x}^{(t)} &= \mathop{\argmax}_{\mathbf{x}} Pr\left(\mathbf{x}|\mathbf{z}^{(1)},\ldots,\mathbf{z}^{(t)}\right) \\
    &=\mathop{\argmax}_{\mathbf{x}}\frac{ Pr\left(\mathbf{x},\mathbf{z}^{(1)},\ldots,\mathbf{z}^{(t)}\right)}{Pr\left(\mathbf{z}^{(1)},\ldots,\mathbf{z}^{(t)}\right)}\\
    &=\mathop{\argmax}_{\mathbf{x}}\frac{Pr\left(\mathbf{z}^{(1)},\ldots,\mathbf{z}^{(t-1)}\right)Pr\left(\mathbf{x}|\mathbf{z}^{(1)},\ldots,\mathbf{z}^{(t-1)}\right)Pr\left(\mathbf{z}^{(t)}|\mathbf{x}\right)}{Pr\left(\mathbf{z}^{(1)},\ldots,\mathbf{z}^{(t)}\right)}\\
    &=\mathop{\argmax}_{\mathbf{x}} Pr\left(\mathbf{x}|\mathbf{z}^{(1)},\ldots,\mathbf{z}^{(t-1)}\right)Pr\left(\mathbf{z}^{(t)}|\mathbf{x}\right).
\end{aligned}
\end{equation}
Combining (\ref{eq3.2.2.1}),(\ref{eq3.2.2.2}) and (\ref{eq3.2.2.3}), we can obtain the closed-form solution of (\ref{eq3.2.2.0}).
In practice, we fix $K = 2$ since it already achieves satisfactory results and larger $K$ may cause both efficiency and overfitting issues.
Moreover, to reflect the fact that $\mathbf{x}^{(t)}$ has a decreasing correlation with $\mathbf{x}^{(t-\tau)}$ when $\tau$ increases, we assume that  

\begin{equation}
\small
\label{eq3.2.2.4}
\begin{aligned}
    \mu_k^{(t)} &= \frac{\sum_{\tau=1}^{t-1}\gamma^{\tau}\mathbf{x}^{(t-\tau)}}{\sum_{\tau=1}^{t-1}\gamma^{\tau}}\ , \\
    \Sigma_k^{(t)} &= \beta_k\Sigma_k + \\ &(1-\beta_k)\frac{\sum_{\tau=1}^{t-1}\gamma^{\tau}\left(\mathbf{x}^{(t-\tau)}-\mu_k^{(t)}\right)^T\left(\mathbf{x}^{(t-\tau)}-\mu_k^{(t)}\right)}{\sum_{\tau=1}^{t-1}\gamma^{\tau}}\ .
\end{aligned}
\end{equation}
where $\gamma, \{\beta_k\}_{k=1}^K \in [0,1]$ along with $2M \times 2M$ positive semi-definite matrices $\{\Sigma_k\}_{k=1}^K$ and $\Sigma_{noise}$ are unknown model parameters.
In practice Eq.~(\ref{eq3.2.2.4}) can be calculated recursively, whose computational complexity remains constant when $t$ increases.

To further reduce the number of parameters, we calculate the covariance matrix $\mathbf{S}$ of all $\mathbf{p}_i^{(t)}-\mathbf{p}_i^{(t-1)}$ in the training set, denote the matrix of eigenvectors of $\mathbf{S}$ as $\mathbf{V}$, and finally assume  
\begin{equation}
\label{eq3.2.2.5}
\begin{split}
    &\Sigma_{noise} = \mathbf{V}^T\Gamma_{noise}\mathbf{V}\ ,\\
    &\Sigma_k = \mathbf{V}^T\Gamma_k\mathbf{V}, k = 1,\ldots,K\ ,
\end{split}
\end{equation}
where $\Gamma_{noise}$ and $\{\Gamma_k\}_{k=1}^K$ are diagonal matrices.
In summary, we have $\Theta = \left[\gamma,\{\alpha_k,\beta_k\}_{k=1}^K,\Gamma_{noise},\{\Gamma_k\}_{k=1}^K\right]$, which are optimized with the loss function in the next subsection. 

\begin{table*}
  \footnotesize
  \caption{Comparisons of NRMSE, AUC and failure rate (at $8.00\%$ NRMSE) on $300$W test set.}
  \label{tab:300W}
  \vspace{-3mm}
  \centering
  \begin{tabular}{l|llllllll}
    \toprule
    Methods    & SDM~(2013) & CFAN~(2014) & CFSS~(2015) & MDM~(2016) & DAN~(2017) & GAN~(2017) & CHR~(2016) & FHR \\
    \midrule
    NRMSE   & $5.83$ & $5.78$ & $5.74$ & $5.05$  & $4.30$   & $3.96$ & $4.07$ & $\bm{3.80}$ \\
    AUC     & $36.3$ & $34.8$ & $36.6$ & $45.3$  & $47.0$   & $53.6$ & $50.9$ & $\bm{55.9}$ \\
    Failure & $13.0$ & $14.0$ & $12.3$ & $6.80$  & $2.67$   & $2.50$ & $2.33$ & $\bm{1.33}$ \\
    \bottomrule
  \end{tabular} \vspace{-0.6mm}
\end{table*}

\begin{figure*}
  \centering
  \subfigure[]{
    \label{fig:STB_td} 
    \includegraphics[width=0.33\textwidth]{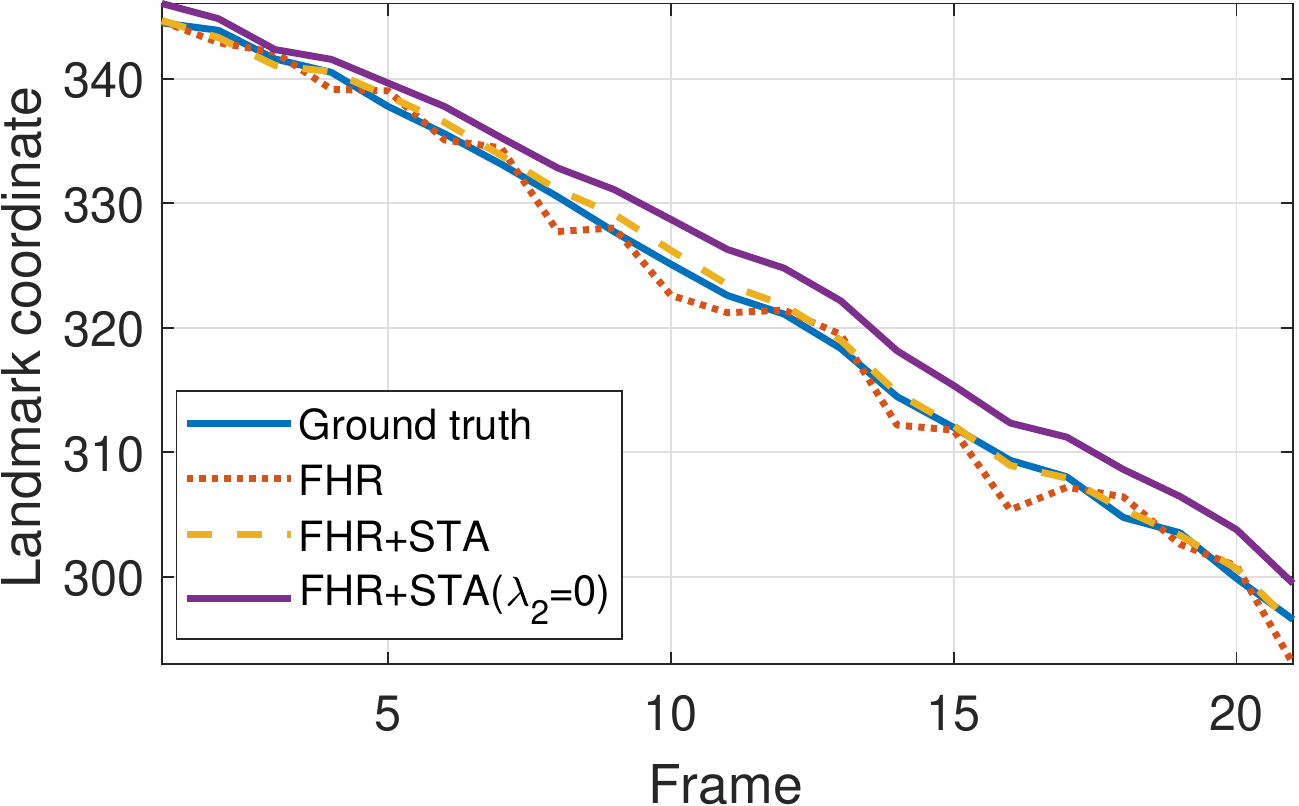}}
  \vspace{1in}
  \hspace{3mm}
  \subfigure[]{
    \label{fig:STB_bb} 
    \includegraphics[width=0.217\textwidth]{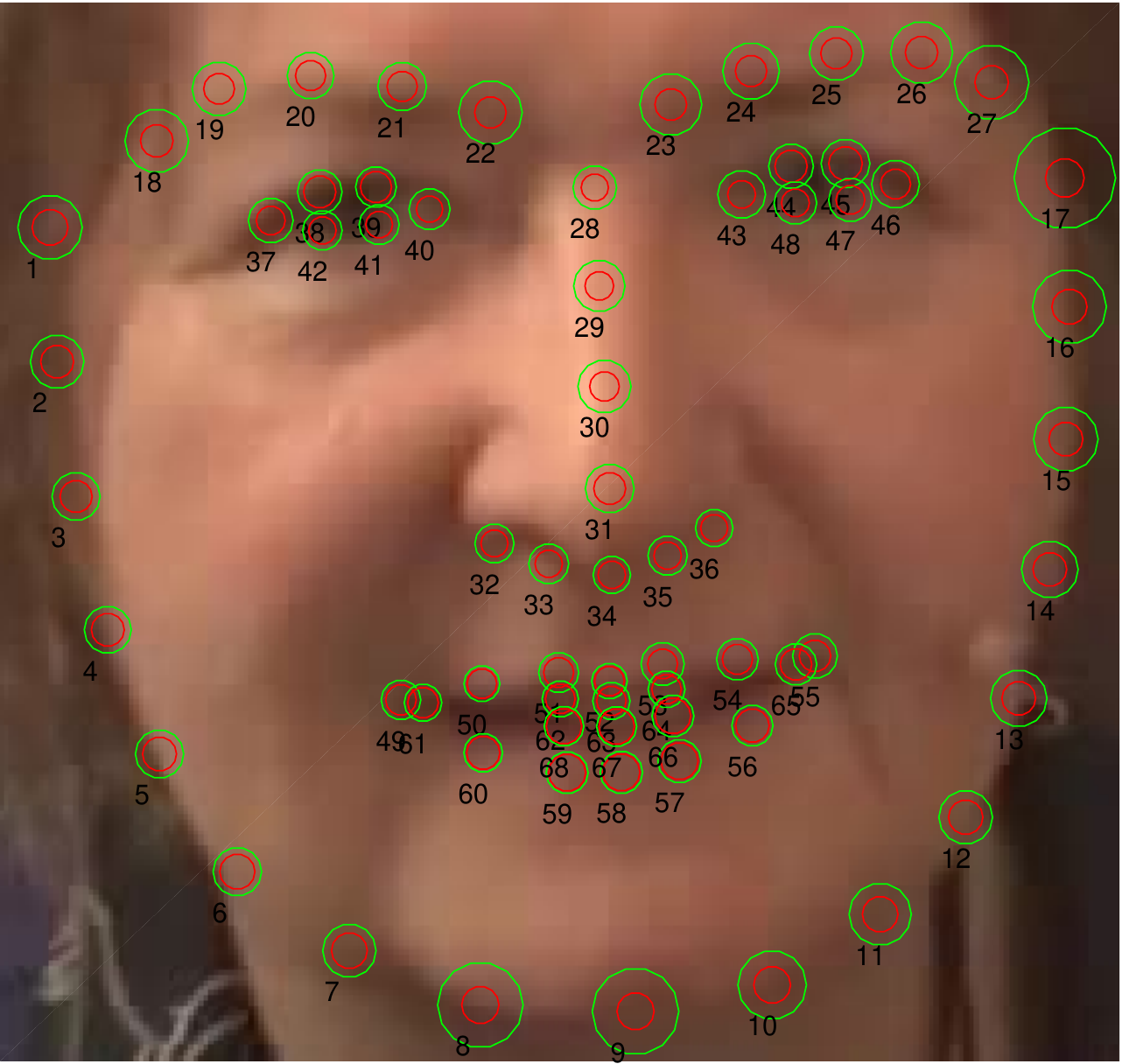}}
  \hspace{3mm}
  \subfigure[]{
    \label{fig:STB_bar} 
    \includegraphics[width=0.35\textwidth]{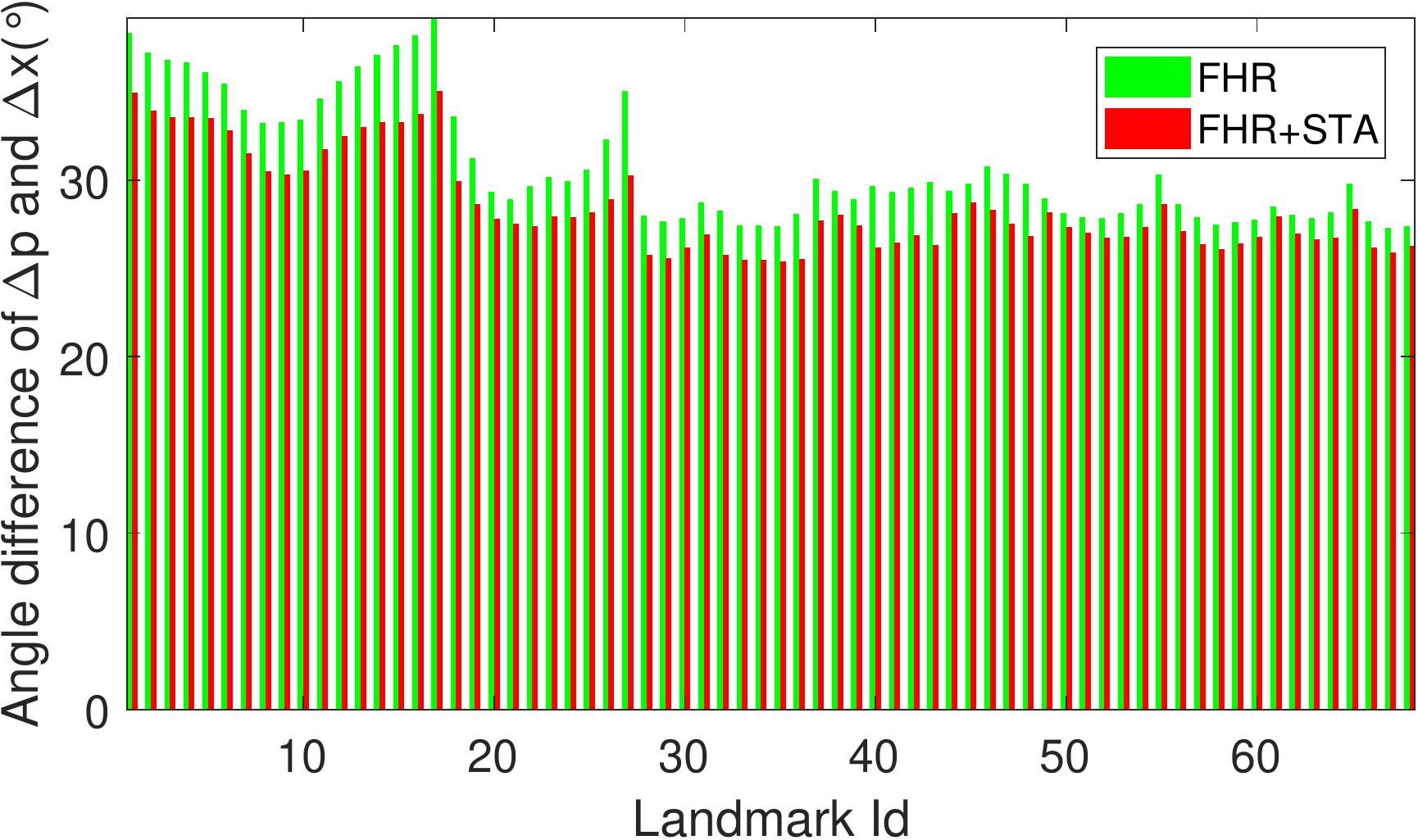}}
  \vspace{-30mm}
  \caption{\footnotesize (a) The effect of stabilization loss for time delay; (b) The magnitute of the stability metric for FHR (green) and FHR+STA (red). (c) The orientation of the stability metric for FHR (green) and FHR+STA (red).}
  \label{fig:ablation_study} 
\end{figure*}

\subsubsection{Loss Function Design} \label{sec:3.2.2}

We now introduce and optimize a novel loss function so as to estimate the above stabilization model parameter $\Theta$.
Throughout this section, we denote all $\mathbf{x}_i^{(t)}$ as $\mathbf{x}_i^{(t)}(\Theta)$ to emphasize that the stabilized landmarks are functions of model parameter $\Theta$.
Our loss function is defined as follows:
\begin{equation}
\label{eq3.2.1.1}
    \mathcal{L}\left(\Theta\right) =  \mathcal{L}_{reg}\left(\Theta\right) + \lambda_1 \mathcal{L}_{tm}\left(\Theta\right).
\end{equation}

This loss has two terms. The first term, $\mathcal{L}_{reg}\left(\Theta\right)$ is called the regularization loss, which regularizes the stabilized output to be close to the ground truth, and is defined as follows:
\begin{equation}
\label{eq3.2.1.2}
   \resizebox{\hsize}{!} {$\mathcal{L}_{reg}\left(\Theta\right) = \frac{\sum_{i=1}^V\sum_{t=1}^{T_i}\|\mathbf{x}_i^{(t)}(\Theta)-\mathbf{p}_i^{(t)}\|_2^2}{\sum_{i=1}^VT_i}+\lambda_2\frac{\sum_{i=1}^V\sum_{t=2}^{T_i}\left(\left(\mathbf{p}_i^{(t)}-\mathbf{p}_i^{(t-1)}\right)^+\left(\mathbf{x}_i^{(t)}(\Theta)-\mathbf{p}_i^{(t)}\right)\right)^2}{\sum_{i=1}^V(T_i-1)}$},
\end{equation}
where $\mathbf{x}^+$ indicates the Moore-Penrose general inverse of vector/matrix $\mathbf{x}$.
We can see that the first term of (\ref{eq3.2.1.2}) is the average Euclidean distance of the ground truth and the model output.
The second term aims to fit every $\mathbf{x}_i^{(t)}(\Theta)$ in terms of $\alpha \mathbf{p}_i^{(t-1)}+(1-\alpha)\mathbf{p}_i^{(t)}$, where $\alpha$ is the coefficient, and estimate the expectation of $\alpha^2$.
If this expectation is large, it means that 1) $\mathbf{x}_i^{(t)}(\Theta)$ is more similar to $\mathbf{p}_i^{(t-1)}$ than $\mathbf{p}_i^{(t)}$, and 2) the model output has a significant time delay, which is undesirable.
Since our stabilization model uses the alignment results of the past frames, how to avoid time delay is a critical task.
Therefore we emphasize the time delay loss as an individual term in $\mathcal{L}_{reg}$ (see Fig.~\ref{fig:SHN_FSHN:a}).

The second term in (\ref{eq3.2.1.1}), $\mathcal{L}_{tm}\left(\Theta\right)$ is called the smooth loss which favors the stabilized output to be smooth, and is defined as follows:
\begin{equation}
\label{eq3.2.1.3}
       \resizebox{\hsize}{!} {$\mathcal{L}_{tm}\left(\Theta\right)=\min_{\bm{q}}\frac{\sum_{i=1}^V\sum_{t=2}^{T_i-1}\left(\|\mathbf{x}_i^{(t)}(\Theta)-q_i^{(t)}\mathbf{x}_i^{(t-1)}(\Theta)-(1-q_i^{(t)})\mathbf{x}_i^{(t+1)}(\Theta)\|_2^2+\lambda_3\|q_i^{(t)}-.5\|_2^2\right)}{\sum_{i=1}^V(T_i-2)}$},
\end{equation}
where $q_i^{(t)} \in \mathbb{R}$ and $\bm{q} = (q_1^{(2)},\ldots,q_V^{(T_V-1)})$. $\mathcal{L}_{tm}\left(\Theta\right)$ can be seen as a trade-off between two stability losses, controlled by $\lambda_3$.
When $\lambda_3 = 0$, it is equivalent to
\begin{equation}
\label{eq3.2.1.4}
    \resizebox{\hsize}{!}
    {$\min_{\bm{q}}\frac{\sum_{i=1}^V\sum_{t=2}^{T_i-1}\|\mathbf{x}_i^{(t)}\left(\Theta\right)-q_i^{(t)}\mathbf{x}_i^{(t-1)}\left(\Theta\right)-(1-q_i^{(t)})\mathbf{x}_i^{(t+1)}\left(\Theta\right)\|_2^2}{\sum_{i=1}^V(T_i-2)} $},
\end{equation}
which is the average distance from $\mathbf{x}_i^{(t)}$ to the line $\overline{\mathbf{x}_i^{(t-1)}\mathbf{x}_i^{(t+1)}}$.

On the other hand, when $\lambda_3 \rightarrow \infty$, $\mathcal{L}_{tm}\left(\Theta\right)$ is equivalent to
\begin{equation}
\label{eq3.2.1.5}
    \resizebox{\hsize}{!}
    {$\frac{\sum_{i=1}^V\sum_{t=2}^{T_i-1}\|\mathbf{x}_i^{(t)}\left(\Theta\right)-\frac{1}{2}\left(\mathbf{x}_i^{(t-1)}\left(\Theta\right)+\mathbf{x}_i^{(t+1)}\left(\Theta\right)\right)\|_2^2}{\sum_{i=1}^V(T_i-2)} $},
\end{equation}
which is the average distance from $\mathbf{x}_i^{(t)}$ to the midpoint of the line $\overline{\mathbf{x}_i^{(t-1)}\mathbf{x}_i^{(t+1)}}$.
The trade-off smooth loss $\mathcal{L}_{tm}(\Theta)$ will cause the loss contour of $\mathbf{x}_i^{(t)}$ to have an ellipse with long axis $\overline{\mathbf{x}_i^{(t-1)}\mathbf{x}_i^{(t+1)}}$ (see Fig.~\ref{fig:SHN_FSHN:b}), which we argue is a more reasonable indicator of the smoothness of the movement trajectory $\overrightarrow{\mathbf{x}_i^{(t-1)}\mathbf{x}_i^{(t)}\mathbf{x}_i^{(t+1)}}$.

With the stabilization model and the loss function, we can estimate the model parameters using the standard optimization method~\cite{Lagarias1998opt}.
The proposed stabilization algorithm is trained by various videos, which learns the statistics of different kinds of movements and thus is more robust than the traditional stabilization model~\cite{Cao2014DDE}.
The optimization process converges within $12$ hours on a conventional laptop.


\section{Experiments} \label{sec:4}

\subsection{Experimental Setup} \label{sec:4.1}

\paragraph{Datasets}
We conduct extensive experiments on both image and video-based alignment datasets, including $300$W~\cite{Sagonas2013align}, $300$-VW~\cite{Shen2015fa300vw} and Talking Face (TF)~\cite{TF2014}.
To test on $600$ images of $300$W private set, we follow~\cite{chen2017adversarial} to use use $3,148$ training images from LFPW, HELEN and AFW datasets. 
To test on $300$-VW, we follow~\cite{Shen2015fa300vw} to use $50$ videos for training and the rest $64$ videos for testing.
Specifically, the $64$ videos are divided to three categories: well-lit (Scenario$1$), mild unconstrained (Scenario$2$) and challenging (Scenario$3$) according to the difficulties.
To test on $5,000$ frames of TF dataset, we follow~\cite{Liu2017TSTN} to use the model trained by the training set of $300$-VW dataset.

\begin{table}
      \footnotesize
      \centering
      \tabcaption{RMSE comparisons on $5$ face scales on $300$W.}
      \vspace{-3mm}
	  \begin{tabular}{c|c|c}
		\toprule
		Face scale & Ave. inter-ocular dis. & CHR/FHR (RMSE~$\downarrow$) \\
		\midrule
		Very small & $61.45$  &  $2.66$/$\bm{2.42}$ ($0.24$~$\downarrow$)  \\
		Small & $91.50$ & $3.71$/$\bm{3.35}$ ($0.36$~$\downarrow$)\\
		Medium & $127.68$  & $5.06$/$\bm{4.55}$ ($0.51$~$\downarrow$)\\
		Large & $179.55$  & $7.59$/$\bm{7.11}$ ($0.48$~$\downarrow$)\\
		Very large & $296.28$ & $11.67$/$\bm{10.67}$ ($\textcolor{red}{\bm{1.00}}$~$\downarrow$)\\
		\bottomrule
	  \end{tabular}
      \label{table:RMSE_scale}
\end{table}

\vspace{-4mm}
\paragraph{Training Setting}
Training faces are cropped using the detection bounding boxes, and scaled to $256\times 256$ pixels.
Following~\cite{chen2017adversarial}, we augment the data (e.g., scaling, rotation) for more robustness to different face boxes.
We use the stacked hourglass network~\cite{Newell2016SHN,chen2017adversarial} as the alignment model.
The network starts with a $7\times 7$ convolutional layer with stride $2$ to reduce the resolution to $128\times 128$, followed by stacking $4$ hourglass modules.
For evaluation, we adopt the standard Normalized Root Mean Squared Error (NRMSE), Area-under-the-Curve (AUC) and the
failure rate (at $8.00\%$ NRMSE) to measure the accuracy, and use the consistency between the movement of landmarks and ground truth as the metric to measure the stability.
We train the network with the Torch7 toolbox~\cite{Torch7}, using the RMSprop algorithm with an initial learning rate of $2.5\times 10^{-4}$, a mini-batch size of $6$ and $\sigma=3$.
Training a fractional heatmap based hourglass model on $300$W takes $\sim$$7$ hours on a P$100$ GPU.

During the stabilization training, we set $\lambda_1 = \lambda_3 = 1$ and $\lambda_2 = 10$ to make all terms in the stabilization loss (\ref{eq3.2.1.1}) on the same order of magnitude.
We estimate the average variance $\rho$ of $\mathbf{z}_i^{(t)}-\mathbf{p}_i^{(t)}$ across all training videos and all landmarks, and empirically set the initial value of $\Gamma_{noise}$ as $\rho\mathbb{I}$.
Also, we initialize $\Gamma_1$ as a zero matrix $\mathbb{O}_{2M\times 2M}$, $\Gamma_2$ as $10\rho\mathbb{I}$, and $\gamma = \beta_1 = \beta_2 = 0.5$.

\begin{figure*}[htb]
\begin{minipage}[b]{0.3\linewidth}
  \centering
  \subfigure{
    \label{fig:Gra_GP} 
    \includegraphics[width=0.72\textwidth]{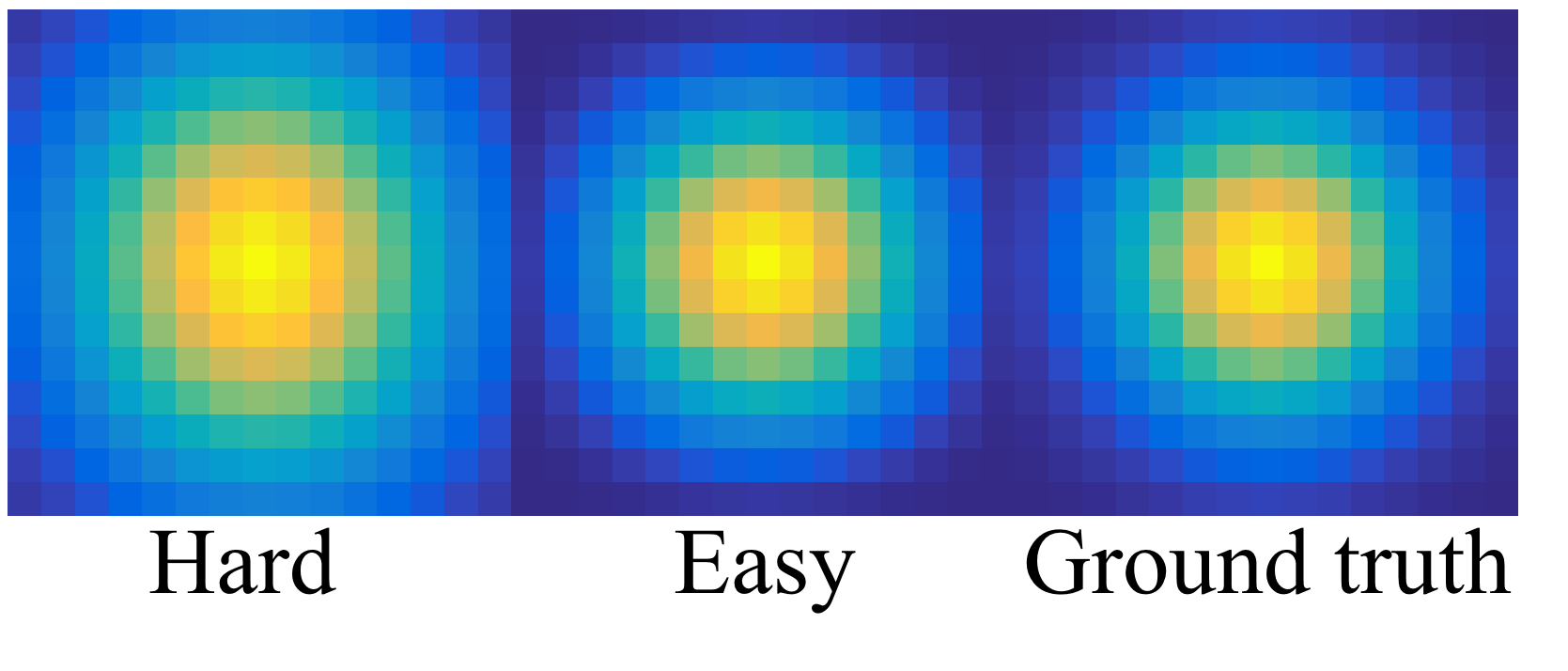}}
  \vspace{-6mm}
  \caption{\footnotesize {Averaged heatmap distributions}.
  }
  \label{fig:UGD} 
  \vspace{-9.5mm}
\end{minipage}%
\hspace{-1mm}
    \begin{minipage}[h]{0.7\linewidth}
      \tiny
      \centering
      \tabcaption{\footnotesize NRMSE/stability comparisons on $300$-VW test set using $68$ landmarks.}
      \vspace{-2mm}
      \begin{tabular}{l|lllll}
        \toprule
        Methods    & FHR & FHR+STA & FHR+STA($\lambda_2=0$) & FHR+STA($\lambda_3=\infty$) & FHR+STA($\lambda_2=0,\lambda_3=\infty$)\\
        \midrule
        Scenario$1$   & $5.07$/$2.79$ & $4.42$/$1.67$ & $5.55$/$\bm{1.64}$ & $\bm{4.40}$/$1.78$ & $4.49$/$1.68$    \\
        Scenario$2$   & $4.34$/$1.85$ & $4.18$/$\bm{1.15}$ & $4.74$/$1.16$ & $\bm{4.16}$/$1.19$ & $4.33$/$1.17$    \\
        Scenario$3$   & $7.36$/$4.48$ & $5.98$/$2.74$ & $7.58$/$\bm{2.57}$ & $\bm{5.96}$/$2.86$ & $6.74$/$2.76$    \\
        \bottomrule
      \end{tabular}
      \label{tab:stablation}
    \end{minipage}
\end{figure*}

\subsection{Ablation Study} \label{sec:4.2}

\paragraph{Fractional vs.~Conventional Heatmap Regression}
We first compare our fractional heatmap regression with the conventional version~\cite{Newell2016SHN} and other state-of-the-art models~\cite{xiong2013supervised,zhang2014coarse,Zhu2015align,trigeorgis2016mnemonic,kowalski2017deep,chen2017adversarial} on $300$W test set.
The training set includes $3,148$ images, and the test set contains $600$ images.
Note that the stacked hourglass networks for our fractional method and the conventional one are the {\it same}.
The only difference is the transformation between the heatmaps and the coordinates, where our method preserves the fractional part.
As shown in Tab.~\ref{tab:300W}, 
our method significantly outperforms the conventional version with an improvement of $0.27\%$ on NRMSE, and is also better than the state-of-the-art model~\cite{chen2017adversarial}.

What's more, the standard NRMSE cannot fully reflect the advantage of our FHR compared to CHR, since it eliminates the scaling effects by dividing the inter-ocular distance.
To further demonstrate the effects of FHR, we calculate RMSE {\it without} normalization.
Specifically, we collect the inter-ocular distances of all $600$ images and evenly divide the distances to five groups w.r.t.~face scales.
Tab.~\ref{table:RMSE_scale} shows that with larger scales, the gap between FHR and CHR is bigger.
Especially in the largest scale, FHR achieves $1$ pixel promotion for each landmark on average.

\vspace{-4mm}
\paragraph{Stabilization Loss for Time Delay}
Here we demonstrate the effectiveness of our proposed time delay term (i.e., the right part of Eq.~\ref{eq3.2.1.2}).
As in Fig.~\ref{fig:STB_td}, compared to the fractional heatmap regression's output, the stabilized output is not only more smooth, but also closer to ground truth landmarks.
Besides, when the time delay term is removed ($\lambda_2 =0$), the stabilized outputs behave lagged behind the ground truth while this phenomena is largely suppressed by using our proposed loss (Eq.~\ref{eq3.2.1.2}).



\begin{figure}[!]
\begin{minipage}[b]{1.07\linewidth}
  \centering
  \subfigure{
    \includegraphics[width=0.51\textwidth]{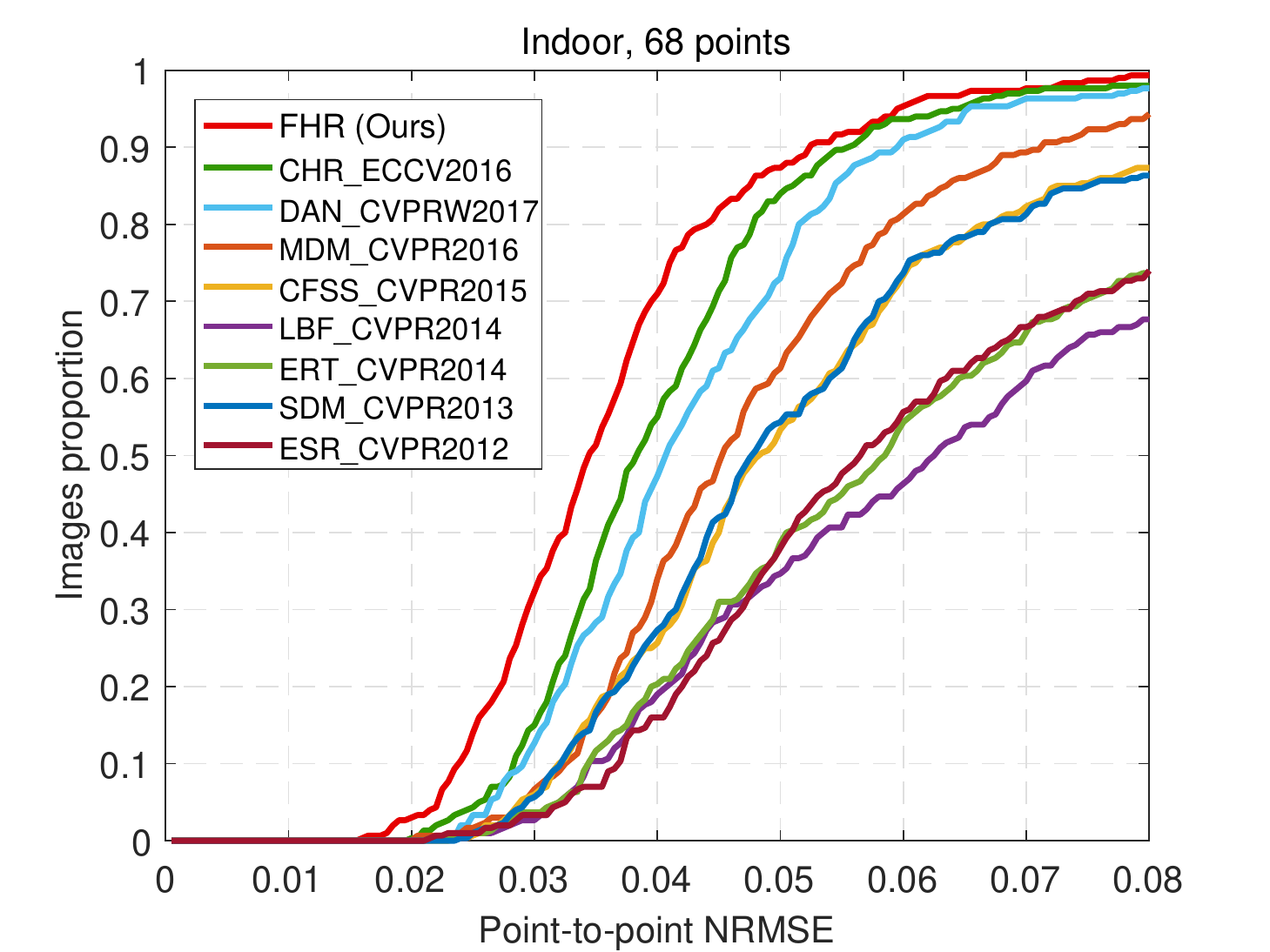}}
  \hspace{-6mm}
  \subfigure{
    \includegraphics[width=0.51\textwidth]{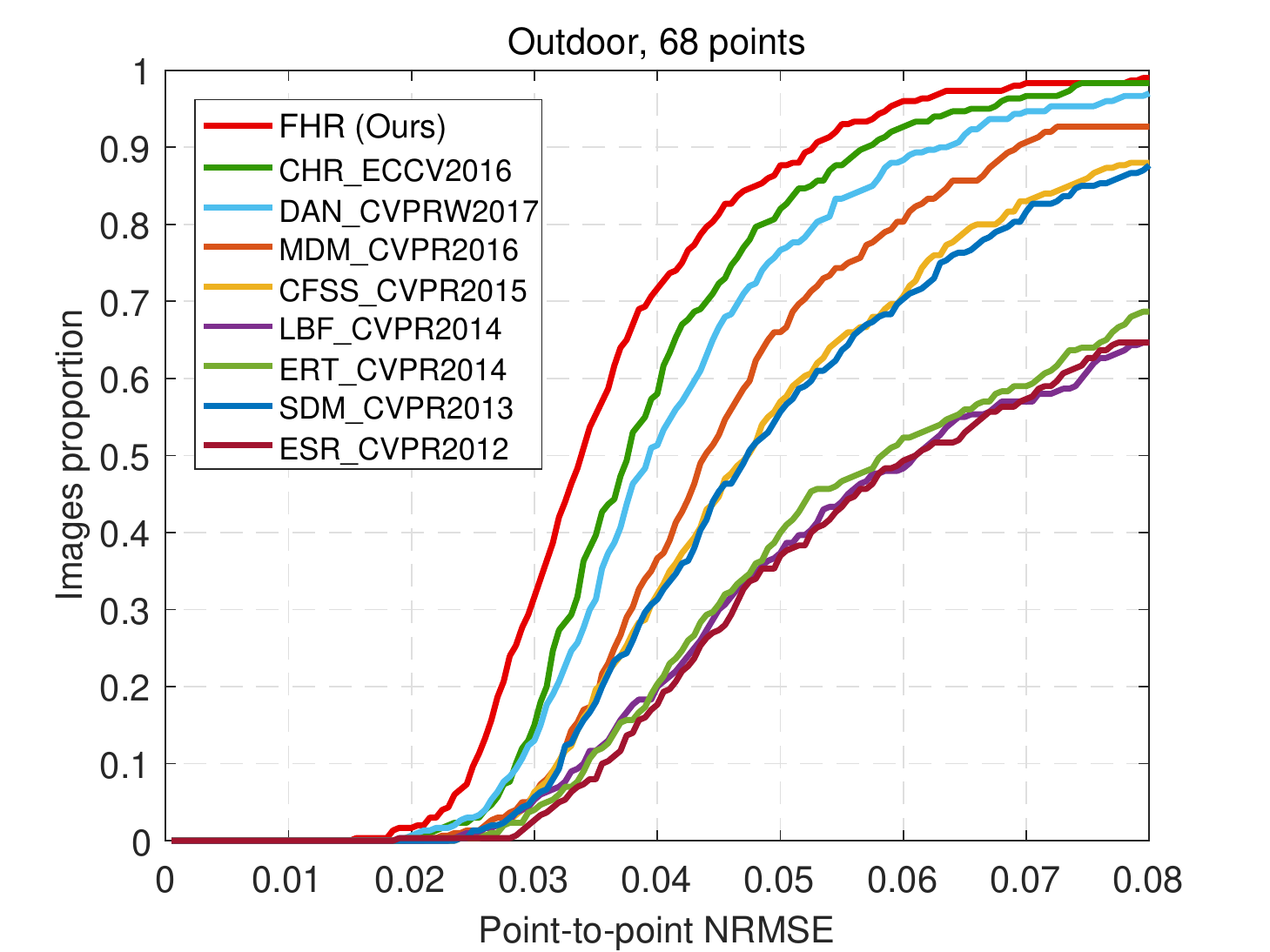}}
  \vspace{-4mm}
  \caption{\footnotesize 
  CED curves on $300$W.
  }
  \label{fig:ced_curves} 
  \vspace{1.4mm}
\end{minipage}%
\end{figure}

\begin{table}[!]
  \footnotesize
  \caption{NRMSE/stability comparisons with REDnet on $300$-VW test set using $7$ landmarks.}
  \label{tab:300vw_7pts}
  \vspace{-3mm}
  \centering
  \begin{tabular}{l|llll}
    \toprule
    Methods    & REDnet~(2016) & FHR & FHR+STA \\
    \midrule
    Scenario$1$   & $8.03$/$10.3$ & $4.44$/$5.49$ & $\bm{3.93}$/$\bm{3.04}$ \\
    Scenario$2$   & $10.1$/$9.64$ & $3.96$/$3.55$ & $\bm{3.82}$/$\bm{3.44}$ \\
    Scenario$3$   & $16.5$/$15.9$ & $5.45$/$4.72$ & $\bm{4.91}$/$\bm{4.45}$ \\
    \bottomrule
  \end{tabular}
\end{table}

\begin{table*} [!]
  \footnotesize
  \caption{NRMSE comparison with state-of-the-art methods on Talking Face dataset using $7$ landmarks.}
  \label{tab:tf_7pts}
  \vspace{-3mm}
  \centering
  \begin{tabular}{l|lllllllllll}
    \toprule
    Methods    & CFAN~(2014) & CFSS~(2015) & IFA~(2014) & REDnet~(2016) & TSTN~(2017) & CHR~(2016) & FHR & FHR+STA  \\
    \midrule
    NRMSE & $3.52$ & $2.36$ & $3.45$ & $3.32$ & $2.13$ & $2.28$ & $\bm{2.06}$ & $2.14$ \\
    \bottomrule
  \end{tabular}
\end{table*}

 \vspace{-4mm}
\paragraph{Stabilization Loss for Smooth}
Next, we evaluate the impact of each term in our loss function (\ref{eq3.2.1.1}) on NRMSE and stability. We use the consistency between the cross-time movement of the landmarks and the ground truth as an indicator of stability.
Specifically, we calculate $\Delta\mathbf{x}_i^{(t)}=\mathbf{x}_i^{(t)}-\mathbf{x}_i^{(t-1)}$ and $\Delta\mathbf{p}_i^{(t)}=\mathbf{p}_i^{(t)}-\mathbf{p}_i^{(t-1)}$ for every video $i$ in the test set, and calculate the average NRMSE between $\Delta\mathbf{x}_i^{(t)}$ and $\Delta\mathbf{p}_i^{(t)}$. Assuming that the ground truth is stable, a lower value indicates higher stability.

The comparison result is shown in Tab.~\ref{tab:stablation}. It can be seen that dropping the time delay term ($\lambda_2 = 0$) causes a higher NRMSE, and changing the smooth loss to the one in~\cite{Cao2014DDE} ($\lambda_3 = \infty$) causes a higher stability loss.
Our proposed method achieves a good balance between accuracy and stability.

\subsection{Comparisons with State of the Arts} \label{sec:4.3}
We now compare our methods FHR and FHR+STA (i.e., the stabilized version) with state-of-the-art models~\cite{Peng2016REDNet,Liu2017TSTN,kowalski2017deep,Zhang2016attr} on two video datasets: $300$-VW and Talking Face.
The comparison adopts two popular settings (i.e., $7$ and $68$ landmarks) used in prior works.

\begin{table*}[t!]
  \footnotesize
  \caption{NRMSE comparison with state-of-the-art methods on $300$-VW test set using $68$ landmarks.}
  \label{tab:300vw_68pts}
  \vspace{-3mm}
  \centering
  \begin{tabular}{l|llllllll}
    \toprule
    Methods       & TSCN~(2016) & CFSS~(2015) & TCDCN~(2016) & TSTN~(2017) & CHR~(2016) & FHR & FHR+STA \\
    \midrule
    Scenario$1$   & $12.5$ & $7.68$ & $7.66$ & $5.36$ & $5.44$ & $5.07$  & $\bm{4.42}$ \\
    Scenario$2$   & $7.25$ & $6.42$ & $6.77$ & $4.51$ & $4.71$ & $4.34$ & $\bm{4.18}$ \\
    Scenario$3$   & $13.10$ & $13.70$ & $15.00$ & $12.80$ & $7.92$ & $7.36$  & $\bm{5.98}$ \\
    \bottomrule
  \end{tabular}
\end{table*}


 \vspace{-4mm}
\paragraph{Comparison with $7$ Landmarks}
First, we evaluate our method on the $300$-VW~\cite{Shen2015fa300vw} dataset and compare with REDnet~\cite{Peng2016REDNet}, using the code released by the authors. 
Tab.~\ref{tab:300vw_7pts} shows the results on the three test sets.
Our proposed method achieves much better performance than REDnet in all cases.
Especially in the hardest Scenario$3$, our method achieves a large improvement of $\sim$$11\%$, which shows the robustness of our method to large facial variations.

Then, we evaluate our method on the Talking Face~\cite{TF2014} dataset compared with state-of-the-art models, such as CFAN~\cite{zhang2014coarse}, CFSS~\cite{Zhu2015align}, IFA~\cite{Asthana2014ifa}, REDnet~\cite{Peng2016REDNet} and TSTN~\cite{Liu2017TSTN}.
Although the annotations of the TF dataset has the same landmark number as $300$-VW dataset, the definitions of landmarks are different.
Therefore, following the setting in~\cite{Liu2017TSTN,Peng2016REDNet}, we use $7$ landmarks for fair comparisons.
The results are shown in Tab.~\ref{tab:tf_7pts}, in which the performance of~\cite{zhang2014coarse,Zhu2015align,Asthana2014ifa} are directly cited from~\cite{Peng2016REDNet,Liu2017TSTN}.
Since the images in TF set are collected in controlled environment and with small facial variations, all of the methods achieve relatively small errors, and our proposed method is still the best.

\vspace{-4mm}
\paragraph{Comparison with $68$ Landmarks}
Next, we evaluate our method on the $300$-VW~\cite{Shen2015fa300vw} dataset under the setting with $68$ landmarks.
The comparison methods include TSCN~\cite{Simonyan2016tsc}, CFSS~\cite{Zhu2015align}, TCDCN~\cite{Zhang2016attr} and TSTN~\cite{Liu2017TSTN}.
We cite the results of~\cite{Simonyan2016tsc,Zhu2015align,Zhang2016attr} from~\cite{Liu2017TSTN}, and list the performance in Tab.~\ref{tab:300vw_68pts}.
As we can see, our proposed FHR achieves the best NRMSEs in all scenarios, and our stabilized version FHR+STA further improves the performance, especially in Scenario$3$.

We then illustrate the Cumulative Errors Distribution (CED) curves of FHR, CHR and some SOTA methods on $300$W in Fig.~\ref{fig:ced_curves}, where the gap between FHR and CHR is competitive to those gaps in prior top-tier works (e.g., CFSS$\&$MDM).
Fig.~\ref{fig:ced_curves} also shows that our FHR contributes more to those relatively easy samples, which makes sense since the insight of FHR is to find a more precise location near a coarse but correct coordinate, whose heatmap output may accurately model the distribution of Ground Truth (GT).
To demonstrate this, we collect some predicted heatmaps from those hard and easy samples, and show their averaged heatmap distributions in Fig.~\ref{fig:UGD} by fixing the centers at the same position respectively.
The easy samples' heatmaps better resemble the Gaussian distribution in GT where FHR can improve the most, while hard samples resemble less and thus FHR contributes little.

\begin{figure}[tbp]
\centering
\tiny
\includegraphics[width=1\linewidth]{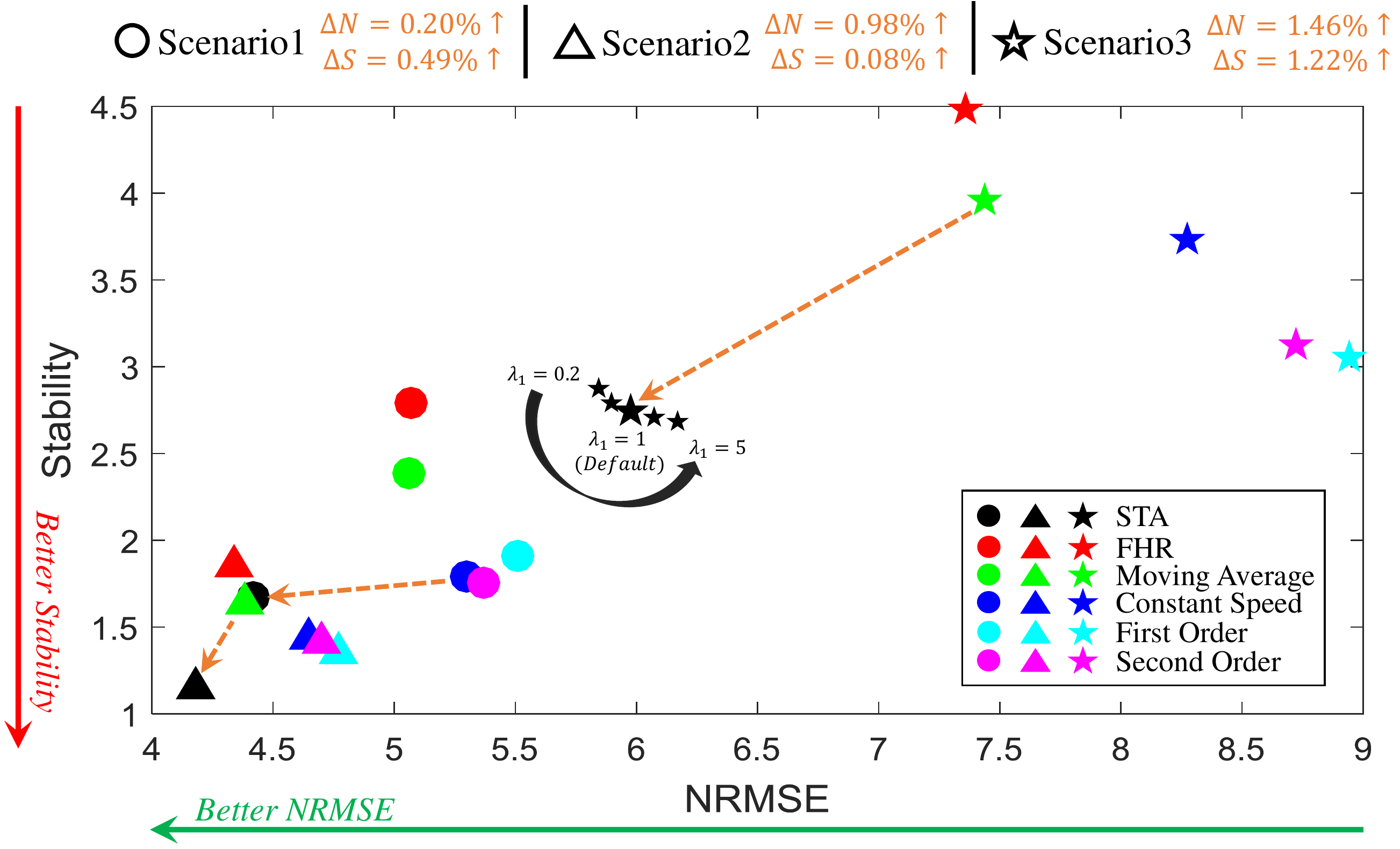}
\vspace{-3mm}
\caption{\small NRMSE/Stability comparisons with four baselines on $300$-VW.
\textit{Dashed lines} indicate the difference between our STA method and the \textit{closest} competitor, where $\Delta N$, $\Delta S$ represent NRMSE, stability improvements, respectively.
\textit{Black arrow and pentagrams} further illustrate the parameter sensitivity of $\lambda_1$, varying among [$0.2$, $0.5$, $1$, $2$, $5$]), on Scenario$3$, which indicates $\lambda_1$ moderately adjusts our stabilization model between accuracy and stability.}
\label{fig:sta_comp3}
\end{figure}

\vspace{-3mm}
\paragraph{Comparison on Stabilization}
We further compare stabilization between our method and REDnet~\cite{Peng2016REDNet} on $300$-VW~\cite{Shen2015fa300vw} with $7$ landmarks.
As shown in Tab.~\ref{tab:300vw_7pts}, our FHR is much more stable than REDnet according to the metric mentioned in~\hyperref[sec:4.2]{Ablation Study}.
To visualize stabilization improvement, we compute  $E (||(\Delta \mathbf{x}-\Delta \mathbf{p})||_2^2)$ for each landmark estimated by FHR and our proposed FHR+STA, and plot them in Fig.~\ref{fig:STB_bb}.
Fig.~\ref{fig:STB_bar} plots the difference of the stability orientation  $E (|<\Delta \mathbf{x}-<\Delta \mathbf{p}|)$ of two methods, where $<$ indicates the orientation of a vector.
We provide videos in our project website, which can effectively demonstrate the stability and superiority of our method.
In some continuous frames, our stabilized landmarks are \textit{even more stable than the ground truth annotations}.

\begin{figure}[!]
\begin{minipage}[b]{1.0\linewidth}
  \centering
  \subfigure{
    \includegraphics[width=0.48\textwidth]{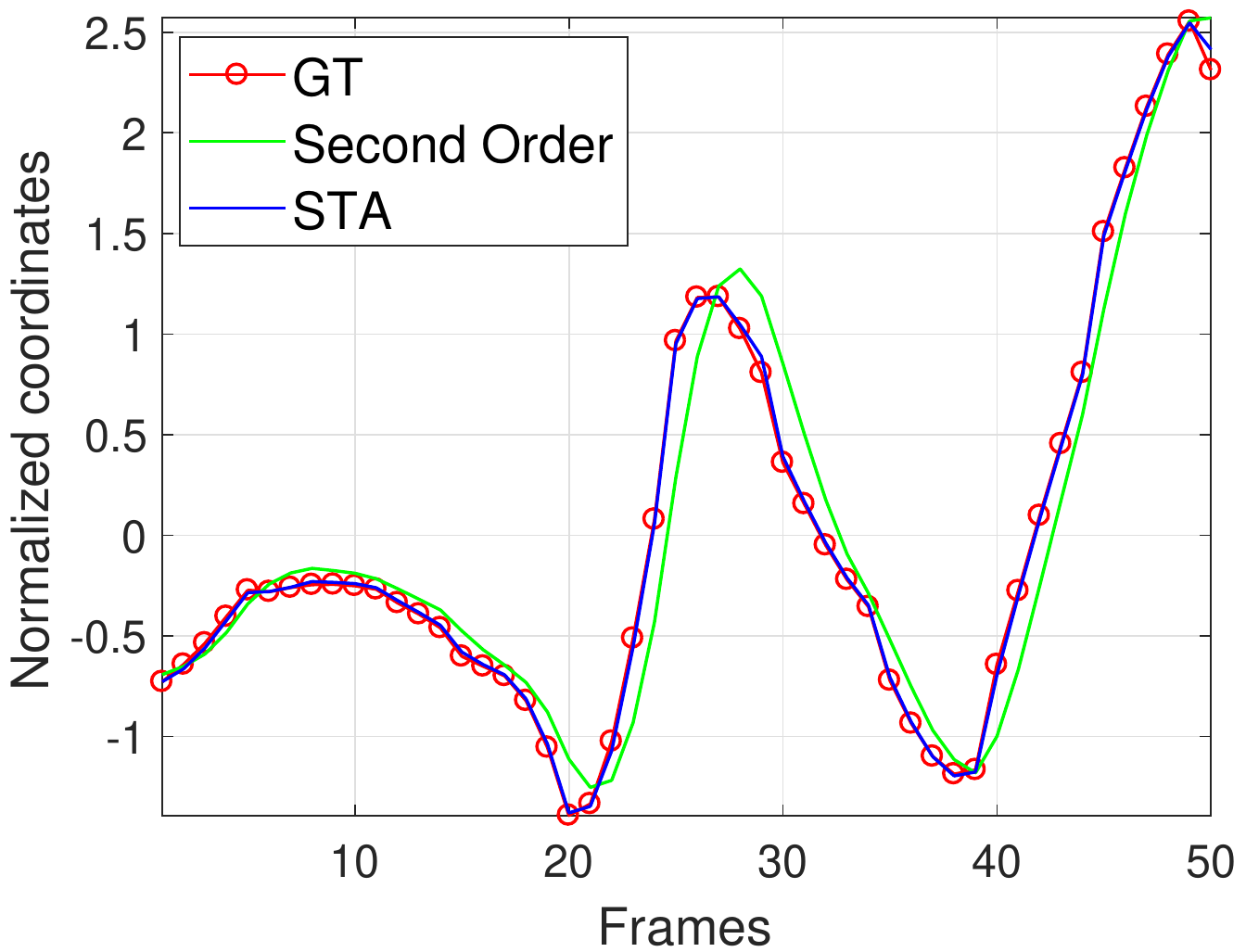}}
  \subfigure{
    \includegraphics[width=0.48\textwidth]{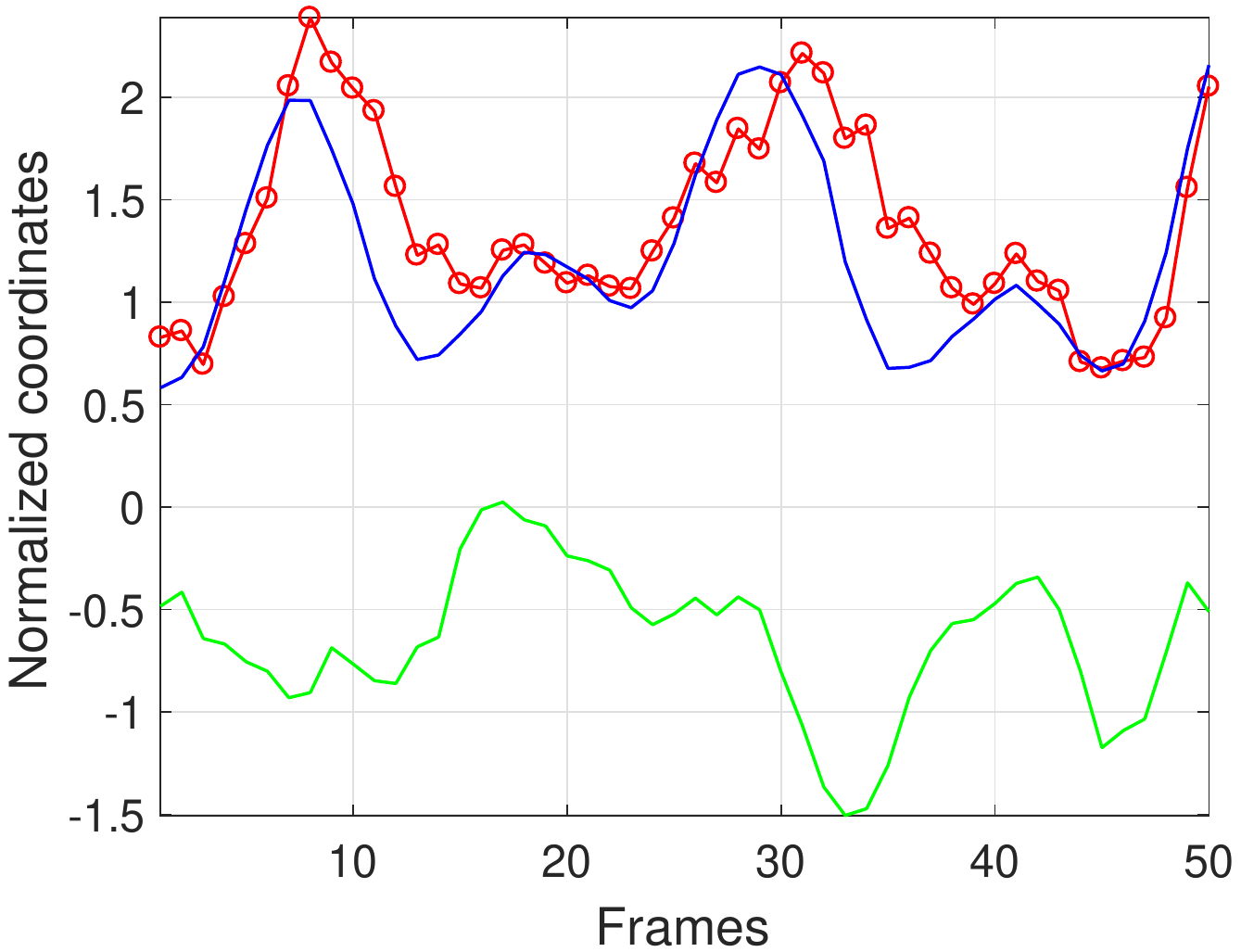}}
  \vspace{-4mm}
  \caption{\footnotesize 
  Stability comparisons with second order stabilization method for time delay (left) and complex movements (right) issues.
  }
  \label{fig:sta_comp_curves} 
  \vspace{1.4mm}
\end{minipage}%
\end{figure}

In addition, Fig.~\ref{fig:sta_comp3} shows that our stabilization model (i.e., STA) significantly outperforms other four baselines in \textit{both of NRMSE and stability}, where all $5$ methods take the same $\mathbf{z}$ from FHR as the input.
Especially in the most challenging set Scenario$3$ where lots of complex movements exist, our method is much better than the \textit{closest} competitor (i.e., moving average), with the improvements of NRMSE and Stability to be $1.46\%$ and $1.22\%$, respectively.
The reasons include: $1$) moving average filter and first order smoother may cause serious time delay problem; $2$) although second order and constant speed methods can handle time delay, it cannot handle multiple movement types (e.g., blinking and turning head).
In contrast, our algorithm can effectively address time delay issue and multiple movement types through the Gaussian mixture setting, and hence is more \textit{precise and stable}.
Fig.~\ref{fig:sta_comp_curves} shows the stability comparisons with the second order stabilization method, which is chosen as a good baseline considering its stability performance.
As we can see, our method significantly outperforms the second order method when handling complex movements, and also shows better ability for time delay issue which is very close to the GT.

\vspace{-3mm}
\paragraph{Time Complexity}
Note that our fractional heatmap regression \textit{does not} impose any additional complexity burden during training compared to the conventional heatmap regression.
For inference, our method provides a closed-form solution to estimate the coordinates from the heatmaps as in Eq.~(\ref{eq3.1.3}), whose runtime is negligible.
Besides, after the parameter $\Theta$ of our stabilization model is learnt, our stabilization algorithm costs $\sim$$6$s to process the entire $64$ test videos of $300$-VW, which costs $5$$\times$$10^{-5}$s per image, and can also be ignored.

\vspace{-2mm}
\section{Conclusions} \label{sec:5}

In this paper, a novel Fractional Heatmap Regression (FHR) is proposed for high-resolution video-based face alignment.
The main contribution in FHR is that we leverage $2$D Gaussian generation prior to accurately estimate the fraction part of coordinates, which is ignored in conventional heatmap regression based methods.
To further stabilize the landmarks among video frames, we propose a novel stabilization model that addresses the time-delay and non-smooth issues.
Extensive experiments on popular benchmarks demonstrate our proposed method is more accurate and stable than the state of the arts. 
Except for the facial landmark estimation task, the proposed FHR has the potential to be plugged into any existing heatmap based system (e.g., human pose estimation task) and boost the accuracy.



{\small
\bibliographystyle{aaai}
\bibliography{egbib}
}

\end{document}